# Image Inpainting by Kriging Interpolation Technique

Firas A. Jassim
Faculty of Administrative Sciences
Management Information Systems Department
Irbid National University
Irbid 2600, Jordan

*Abstract*—Image inpainting is the art of predicting damaged regions of an image. The manual way of image inpainting is a time consuming. Therefore, there must be an automatic digital method for image inpainting that recovers the image from the damaged regions. In this paper, a novel statistical image inpainting algorithm based on Kriging interpolation technique was proposed. Kriging technique automatically fills the damaged region in an image using the information available from its surrounding regions in such away that it uses the spatial correlation structure of points inside the k×k block. Kriging has the ability to face the challenge of keeping the structure and texture information as the size of damaged region heighten. Experimental results showed that, Kriging has a high PSNR value when recovering a variety of test images from scratches and text as damaged regions.

## I. INTRODUCTION

The filling-in of missing or unwanted information is an extremely considerable topic in image processing. The most important applications of image inpainting are objects removal, scratch removal, restoring missing areas, image repairing, etc. Actually, an image or photograph is sometimes damaged because of aging. Therefore, the exact definition of inpainting is that the reconstruction of damaged images in such a way that is unnoticeable by the human eye. The manual work of inpainting is most often a very time consuming process. Due to digitalization of this technique, it is automatic and faster. The most essential inpainting technique is the diffusion-based technique [8][9][18]. In these techniques, the missing blocks are filled by diffusing the image pixels from the observed blocks into the missing blocks. These techniques are based on the theory of partial differential equation (PDE). According to [9], the holes are filled by procreating the isophote into the missing blocks. The isophote are lines of equal grey values. Furthermore, a Navier-Strokes equation in fluid dynamics have been utilized into the field of image inpainting [8]. Moreover, Total Variational (TV) inpainting technique was proposed by [18], which uses an Euler-Lagrange equation recovers the missing information. In the Total Variational and inside the inpainting domain the model simply employs anisotropic diffusion based on the contrast of the isophotes to incorporate the principle of continuity. It must be mentioned that, there is a pert for the statistics in the field of image inpainting. This area has been researched by [3][16]. Additionally, [15] has discussed the application of K-nearest neighbour algorithm in image inpainting task. Also, a simple and faster image inpainting algorithm that uses Gauss convolution kernel has been proposed by [10]. The Radial Basis Functions (RBF) for reconstruction of damaged images and for eliminating noises from corrupted images was researched by [13]. The first appearance for exemplar based inpainting method was firstly discussed by [1]. In exemplar based method, the target region is filled with patches from the surrounding area that have similar texture. The process of selecting candidate patches is done with special priority to those along the isophotes. As a modification and addition to exemplar based method, [7] proposed a novel texture formation method called coherence-based local searching (CBLS) for region filling. The basic idea in (CBLS) is that instead researching in the whole source region, a minimization procedure may be implemented on the researching area of patches in the surrounding regions that can outfit adequate information to resolve which region must be filled. Another modification of exemplar based method was proposed by [20], through investigating the sparsity of natural image patches. According to [2], a novel algorithm based on a cellular neural network has been proposed. The diffusion-based inpainting algorithms have achieved convincingly excellent results for filling the non-textured or relatively smaller missing region. However, they tend to introduce smooth effect in the textured region or larger missing region.

In this paper, a novel technique based on Kriging interpolation method for spatial data was proposed. The organization of this paper is as follows: In section II, an unpretentious background concerning Kriging interpolation was presented. The proposed technique was discussed in section III with an illustrative example. In section IV, experimental results have been presented. These results contain ocular and numerical results to support the proposed technique. Finally, the conclusions and inferences were introduced in section V.

## II. KRIGING PRELIMINARIES

Kriging is a geostatistical interpolation method that takes into account both the distance and the degree of variation between known points when predicting values in unknown locations. Kriging is aiming to estimate unknown values at specific points in space by using data values from its surrounding regions. Kriging yields optimal aftermaths compared with the traditional interpolation methods [5]. It must be mentioned that, Kriging is an exact interpolator technique because it ensures that the original observed values will stay as it, i.e. the old values will not affected by the interpolation technique. Kriging predictions are treated as weighted linear combinations of the known locations. According to Kriging technique, the closer the input, the more positively correlated predictions [19]. Now, let's bring the previously mentioned thoughts into digital image processing. According to [14], the pixels within the same k×k block are highly correlated, therefore; the application of Kriging inside the k×k block will yields high positively correlated predictions. Kriging gives weights for each point inside k×k block in accordance to its distance from the unknown value. Actually, these predictions treated as weighted linear combinations of the known values. The weights should provide a Best Linear Unbiased Estimator (BLUE) of the predicted point [11]. The essential characteristic of Kriging over conventional interpolation methods is that it uses the spatial correlation structure of points inside k×k block being interpolated in order to compute the unknown point [12]. There is a robust connection between image denoising and image inpainting especially scratch removal. Both fields are sharing the same principles in finding and removing the unwanted areas [6].

The basic formula of Kriging technique may be represented as follows:

$$\hat{P}^* = \sum_{i=1}^{N} \lambda_i P_i \qquad (1)$$

where N is the total number of the non-scratched pixels insides the k×k block. Moreover, $\hat{P}^*$ stands for the predicted pixel and $P_i$ are the representation for the non-scratched pixels insides the k×k block. The weights of the non-scratched pixels $\lambda_i$ must satisfy:

$$\sum_{i=1}^{N} \lambda_i = 1 \qquad (2)$$

The Kriging estimate is obtained by choosing $\lambda_i$ that minimize variance of the estimator under the unbiasedness constraint:

$$\sigma^2 = E[(P - \hat{P}^*)^2] \qquad (3)$$

There are several Kriging types, differ in their treatments for the weighted components (λ's). The most preferred Kriging type and it is considered to be the best one is ordinary Kriging because it minimizes the variance of the prediction error [19].

It must be mentioned that, variogram is one of the most supporting functions to indicate spatial correlation in observations measured at observed points. The variogram is a function of the distance and direction separating two locations that is used to quantify dependence. The variogram is defined as the variance of the difference between two variables at two locations [12]. The variogram generally increases with distance and is described by nugget, sill, and range parameters. If the data is stationary, then the variogram and the covariance are theoretically related to each other. It is commonly represented as a graph that demonstrates the variance with respect to the distance between all points of the observed locations [4]. The variogram describes the variance of the difference of samples within the data set and is calculated by the following equation:

$$2\gamma(h) = \frac{1}{n} \sum_{i=1}^{N} [P(x_i) - P(x_i + h)]^2 \qquad (4)$$

where $P(x_i)$ and $P(x_i+h)$ are the pixel values at locations x and $x_i+h$, respectively. In this paper, Kriging was treated as a supporting scheme that helps to reach the goal which is image inpainting. Hence, there is no need to discuss the variogram in a detailed manner. An exhaustive discussion and analysis about variogram could be found through recommended readings [4][12].

## III. PROPOSED TECHNIQUE

In this work, a novel image inpainting method based on Kriging interpolation technique was proposed. The proposed method starts with identifying the queer pixels within the k×k block from the contaminated image. The contamination may be thin scratch, thick scratch, text, bad areas generated by aging, or even unwanted objects that may be eliminated from the original image. These contaminated areas will be marked according to its corresponding mask. After that, the k×k block will be dispatched to Kriging interpolation technique to predict the contaminated areas using the accurate prediction feature of Kriging. As mentioned previously, Kriging method uses variogram to express the spatial variation, and it minimizes the error of predicted values which are estimated by spatial distribution of the predicted values. The resulted predictions seem to be very close to the original pixels. Therefore, Kriging is very suitable to estimate the mask's pixels accurately.

### A. Illustrative Kriging Example

The discussion of Kriging combined with the variogram analysis is so lengthy. Therefore, a practical example will be introduced to explain the previous discussion. An arbitrary block was taken from bitmap Lena image, Fig. (1). The block was contaminated with a scratch, Fig. (2). After the implementation of Kriging, the predicted values at the contaminated locations are too close to the original data values, Fig. (3). Clearly, in Fig. (4), the error values are highly acceptable which reflects that the implementation of Kriging interpolation technique produces excellent approximations that are too close to the original observations.

```
123  124  125  115  119  113  121  125  124
123  122  125  120  121  122  125  123  124
121  122  124  126  122  120  127  124  121
121  122  120  124  121  123  125  126  128
121  123  122  123  121  125  133  122  123
122  118  126  127  123  124  121  125  125
123  120  121  129  119  125  123  126  129
125  123  118  121  122  122  123  133  128
```
Figure 1. Random Block from Lena Bitmap Image

```
123  124  125  115  119  113  121  125  124
 0   122  125  120  121  122  125   0    0
 0   122  124  126   0    0    0   124  121
121   0    0    0   121  123  125  126  128
 0    0   122  123  121  125  133  122  123
122  118   0   127  123  124  121  125  125
123  120   0   129  119  125  123  126  129
125  123  118   0   122  122  123  133  128
```
Figure 2. The Block contaminated with scratch

```
123  124  125  115  119  113  121  125  124
121  122  125  120  121  122  125  124  122
119  122  124  126  122  123  125  124  121
121  122  123  123  121  123  125  126  128
120  120  122  123  121  125  133  122  123
122  118  122  127  123  124  121  125  125
123  120  122  129  119  125  123  126  129
125  123  118  122  122  122  123  133  128
```
Figure 3. Implementation of Kriging to predict the scratched points

```
0   0   0   0   0   0   0   0   0
2   0   0   0   0   0   0  -1   2
2   0   0   0   0  -3   2   0   0
0   0  -3   1   0   0   0   0   0
1   3   0   0   0   0   0   0   0
0   0   4   0   0   0   0   0   0
0   0  -1   0   0   0   0   0   0
0   0   0  -1   0   0   0   0   0
```
Figure 4. Error values between the original and the predicted blocks

As a result, Kriging interpolation could be implemented to remove unwanted areas in an image. These unwanted areas may be scratches, text, object, etc. As an example of text removal, Fig. (5) shows a text removal for some selected area while keeping the remaining text as it is. The filling-in process was reached using Kriging interpolation technique with the mask given in Fig. (5,b). The results were highly acceptable because the filling-in process was unnoticeable by human eye which is the main goal of image inpainting.

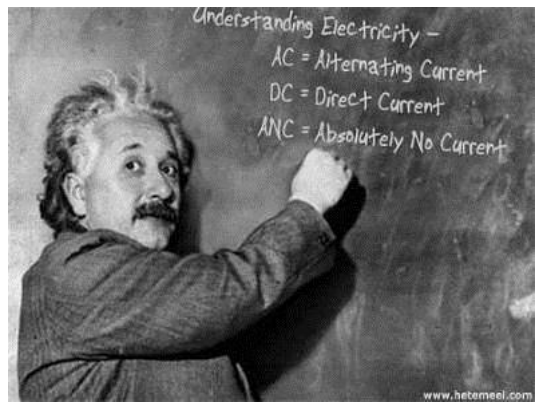
(a)
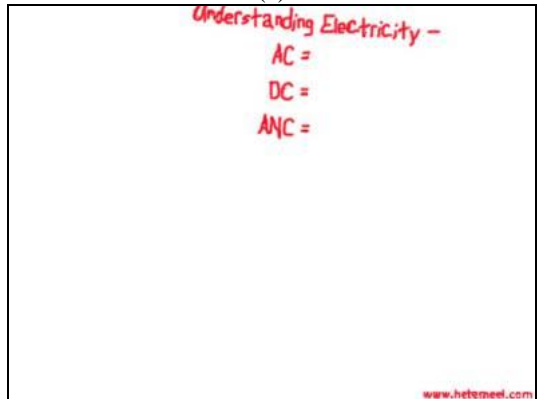
(b)
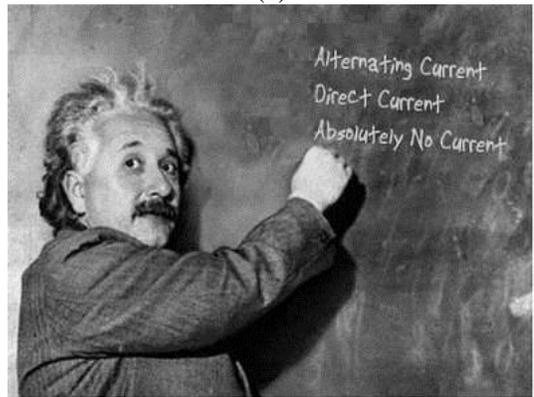
(c)

Figure 5. (a) Einstein Original (b) Mask (c) Selected text removed

IV. EXPERIMENTAL RESULTS

In order to demonstrate the proposed inpainting technique, ten bitmap test images were used as test images, Fig. (6). Additionally, four masks types were implemented to examine the proposed technique. The selection of masks was very adequate such that all kind of masks will be covered, Fig. (7). Starting from Thick scratch, thin scratch, low text, and heavy text, Kriging technique produces very sophisticated results according to the ocular reconstructed images. Furthermore, a standard measure that tests the quality of the reconstructed image is the Peak Signal to Noise Ratio (PSNR) [14]:

$$PSNR = 20\log_{10}\frac{255}{\sqrt{MSE}} \quad (5)$$

where the MSE is shortened for mean square error that calculated as:

$$MSE = \frac{1}{NM}\sum_{x=1}^{N}\sum_{y=1}^{M}[f(x,y)-g(x,y)]^2 \quad (6)$$

According to the calculated PSNR values for the ten test images, table (I), it can be concluded that the results are excellent since it lie within the acceptable range which is 30 to 50 in conformity with [17].

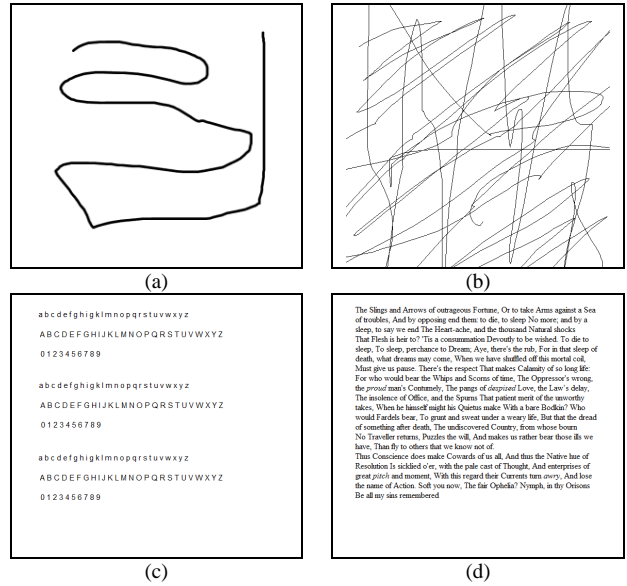

(a)  (b)

(c)  (d)

Figure 7.  Four types of implemented masks

TABLE I.    PSNR VALUES FOR THE 10 TEST IMAGES

|        | Mask 1  | Mask 2  | Mask 3  | Mask 4  |
|--------|---------|---------|---------|---------|
| Lena   | 38.0277 | 45.4131 | 41.3416 | 34.3702 |
| Baboon | 31.3599 | 35.0392 | 33.3432 | 26.7299 |
| Peppers| 39.2157 | 46.4947 | 43.9334 | 35.7265 |
| F16    | 36.0677 | 47.5502 | 40.3283 | 33.4850 |
| Boat   | 34.9864 | 41.1897 | 37.9275 | 29.9704 |
| Mosque | 32.0316 | 36.6053 | 35.7167 | 35.7167 |
| Bird   | 34.8415 | 42.7062 | 39.2466 | 31.9359 |
| Boys   | 33.5322 | 37.4355 | 36.6726 | 29.8040 |
| Saif   | 36.3471 | 41.5247 | 39.3742 | 32.6942 |
| Aws    | 31.7789 | 39.4737 | 36.1827 | 29.5661 |

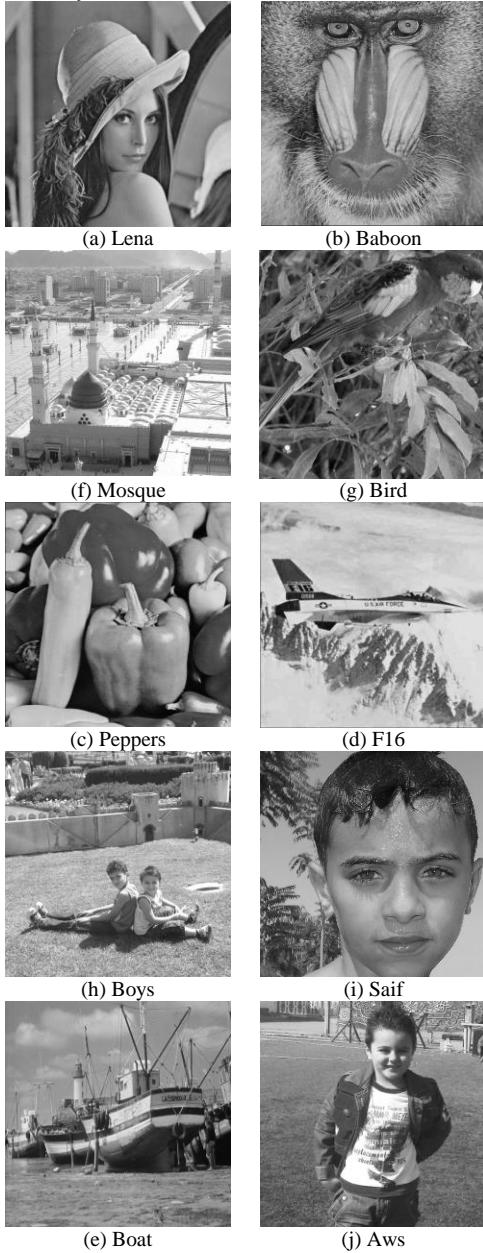

(a) Lena  (b) Baboon
(f) Mosque  (g) Bird
(c) Peppers  (d) F16
(h) Boys  (i) Saif
(e) Boat  (j) Aws

Figure 6.  Ten Test images

V.  CONCLUSIONS

In this paper, a novel approach for removing scratches and text from contaminated images has been presented. The proposed technique use Kriging in a way that removes unwanted regions from image which is known as image inpainting. Despite Kriging being more computationally expensive, it has been shown that it gives very sophisticated output when repairing digital images that have scratches or unwanted text. Experimental results reveal that the proposed Kriging technique having high PSNR value when implemented on a variety of test images.

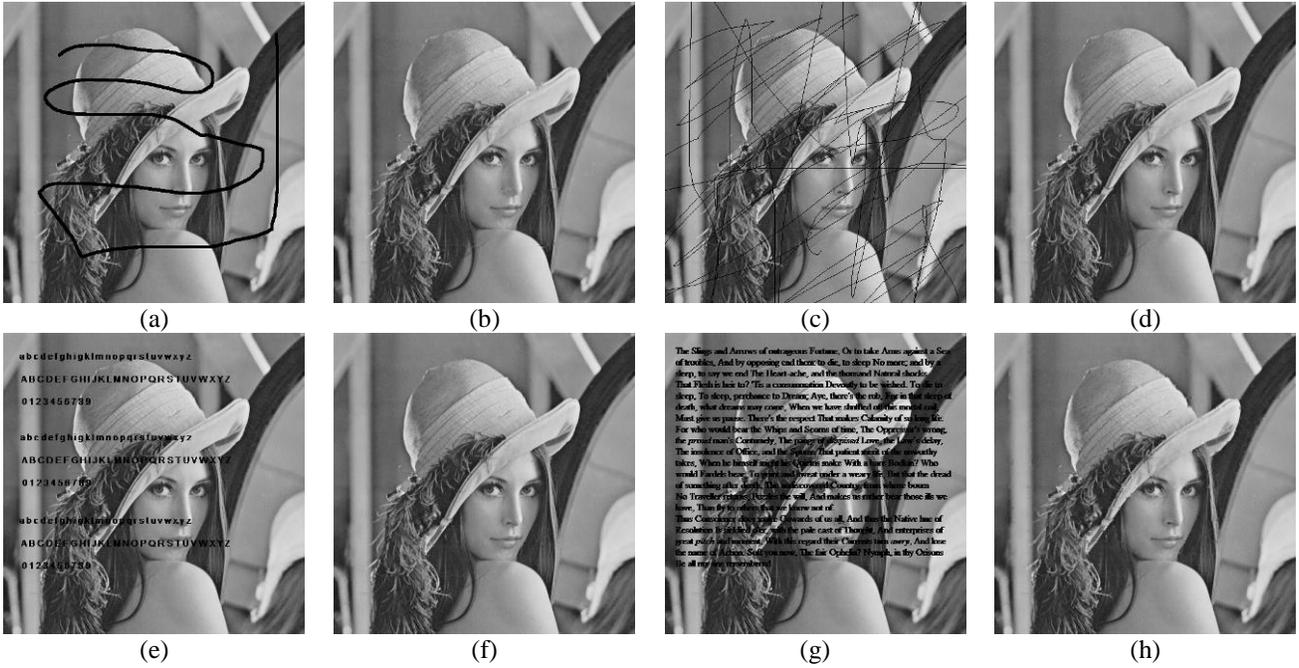

Figure 8. (a) Lena with mask 1 (b) Restored after mask 1 (c) Lena with mask 2 (d) Restored after mask 2 (e) Lena with mask 3 (f) Restored after mask 3

(g) Lena with mask 4 (h) Restored after mask 4

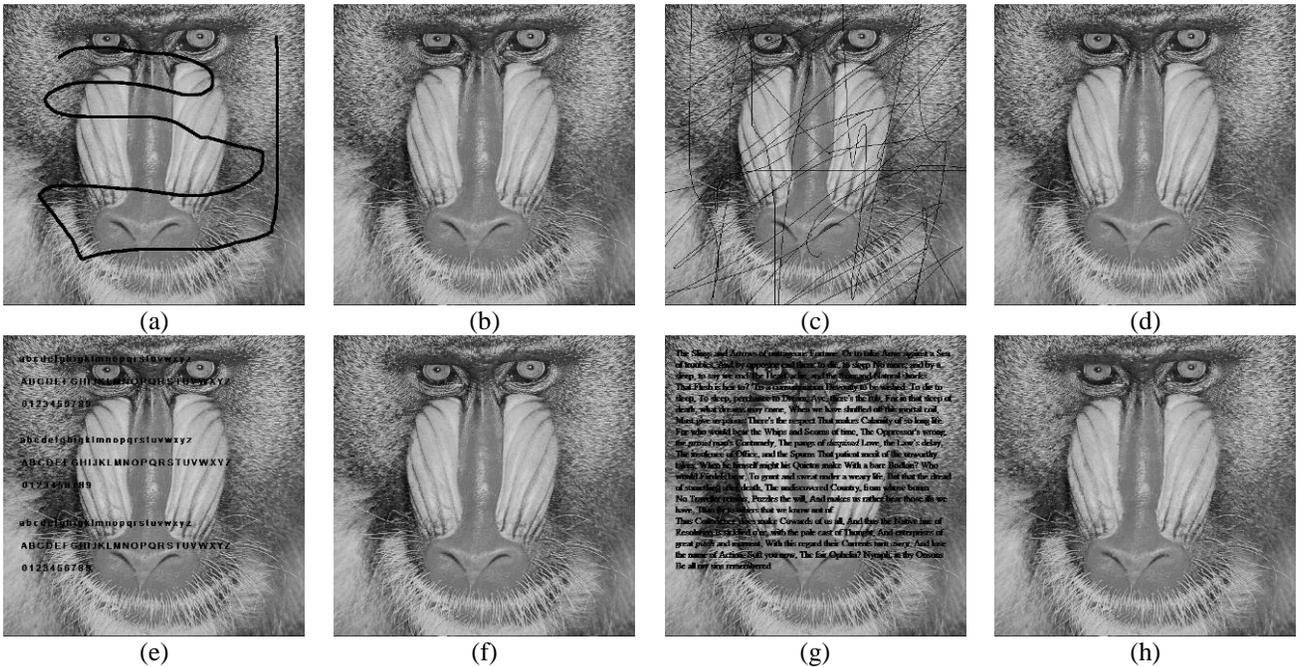

Figure 9. (a) Lena with mask 1 (b) Restored after mask 1 (c) Baboon with mask 2 (d) Restored after mask 2 (e) Baboon with mask 3 (f) Restored after mask 3

(g) Baboon with mask 4 (h) Restored after mask 4

**Firas Ajil Jassim** was born in Baghdad, Iraq, in 1974. He received the B.S. and M.S. degrees in Applied Mathematics and Computer Applications from Al-Nahrain University, Baghdad, Iraq, in 1997 and 1999, respectively, and the Ph.D. degree in Computer Information Systems (CIS) from the Arab University for Banking and Financial Sciences, Amman, Jordan, in 2012. In 2012, he joined the faculty of the Department of Business Administration, Management Information Systems Department, Irbid National University, Irbid, Jordan, where he is currently an assistance professor. His current research interests are image compression, image interpolation, image segmentation, image enhancement, and simulation.